\documentclass[UTF8,11pt,a4paper]{ctexart}

\usepackage[a4paper,margin=2.5cm]{geometry}
\usepackage{amsmath,amssymb}
\usepackage{booktabs}
\usepackage{graphicx}
\usepackage{caption}
\usepackage{subcaption}
\usepackage{multirow}
\usepackage{threeparttable}
\usepackage{array}
\usepackage[hidelinks]{hyperref}
\usepackage{enumitem}
\usepackage[super]{gbt7714}
\begin{document}

% ==================== 中文题名 + 摘要页 ====================
\begin{center}
{\Large\bfseries AI 组件对智慧农业平台绿色价值的边际贡献仿真评估\\
\large 基于“AI能力—农户行为—环境效益”与“IoT—灌溉决策”双链路仿真实验}\\[6pt]
{\normalsize（English Title: Simulating the Marginal Green Contribution of AI Modules in a Smart-Agriculture Platform: Evidence from Two Monte Carlo Experiments）}\\[6pt]
{\normalsize 李肇阳，张锐杰，孙兆吉，张璐\\[2pt]
三亚学院（Sanya University），海南三亚 572022；邯郸未至之境人工智能基础软件有限责任公司}\\[2pt]
{\small 2026年9月}
\end{center}

\begin{quote}
\noindent\textbf{摘要：}智慧农业平台通常将 AI 诊断、物联感知与决策推送捆绑打包，其环境效益的归属难以拆解，导致绿色投入缺乏优化依据。本文在前期“平台整体→减药—减肥—节水—降碳”蒙特卡洛评估的基础上，把平台组件显式化，开展两项受控仿真实验：实验一沿“AI 能力→农户行为→绿色投入减量”链路，将减药/减肥建模为可避免盲目投入占比、处方有效性、决策触达覆盖率与采纳率的乘积，比较经验农技模式与 AI 模式；实验二沿“IoT 传感器+AI 灌溉决策→节水与稻田甲烷”链路，比较现状(P0)、IoT 工程改造(P1)与 AI 决策(P2)三阶段。结果表明：（1）AI 的绿色价值来自“触达放大器”与“质量放大器”的叠加：经验模式达成“农药减量$\geq$20\%”的概率几乎为零，AI 模式为 20.7\%（基准）至 49\%（准确率0.95、采纳率0.85）；化肥减量$\geq$15\% 的达成概率由约0升至52.0\%。（2）综合灌溉节水率中位数由 P0 的 7.8\% 升至 P1 的 11.0\% 与 P2 的 16.0\%，AI 决策相对工程改造再贡献 5.0pp；稻田CH$_4$减排在 AI 调度下达 30.5\%（人工仅19.8\%），水稻“灌溉—甲烷”子系统碳强度下降27.9\%。（3）两实验敏感性一致指向：绿色目标达成的第一瓶颈是农户采纳率而非算法准确率；AI 数据融合对传感观测误差稳健。研究为智慧农业平台“组件级”绿色价值评估与推广运营优化提供了可复现的仿真框架。
\end{quote}

\noindent\textbf{关键词：}智慧农业；AI诊断；物联网灌溉决策；绿色价值分解；蒙特卡洛实验；采纳率

\newpage

% ==================== English full text ====================
\title{Simulating the Marginal Green Contribution of AI Modules in a Smart-Agriculture Platform: Evidence from Two Monte Carlo Experiments}
\author{Zhaoyang Li, Ruijie Zhang, Zhaoji Sun, Lu Zhang\\
\small \textit{Sanya University, Sanya, Hainan 572022, China;}\\
\small \textit{Handan Weizhi Jingjie AI Basic Software Co., Ltd.}}
\date{September 2026}
\maketitle

\begin{abstract}
Smart agriculture platforms usually bundle AI diagnosis, IoT sensing and decision push into a single package, so the green benefit attributable to each component remains unclear and resource-allocation decisions lack quantitative evidence. Building on a previous platform-level Monte Carlo assessment, this paper makes the components explicit and conducts two controlled simulation experiments. Experiment 1 follows the chain ``AI capability $\rightarrow$ farmer behavior $\rightarrow$ agrochemical input reduction'', modeling pesticide/fertilizer reduction as the product of avoidable blind-application share, prescription effectiveness, decision-touch coverage and adoption rate, and compares an experienced-extension mode with the AI mode: the probability of reaching a 20\% pesticide reduction is essentially zero in the extension mode but 20.7\% at baseline and up to 49\% with diagnosis accuracy 0.95 and adoption 0.85 under AI; the probability of a 15\% fertilizer reduction rises from near zero to 52.0\%. Experiment 2 compares current practice (P0), IoT engineering retrofit (P1), and P1 plus AI irrigation scheduling (P2) for water saving and paddy methane: median aggregate irrigation water saving rises from 7.8\% (P0) to 11.0\% (P1) and 16.0\% (P2), with AI adding 5.0 percentage points beyond engineering; paddy CH$_4$ reduction reaches 30.5\% under AI scheduling versus 19.8\% under manual operation, and the carbon intensity of the rice irrigation-methane subsystem declines by 27.9\%. Sensitivity analyses of both experiments consistently indicate that the primary bottleneck for meeting green targets is the farmer adoption rate rather than algorithm accuracy, and that AI data fusion is robust to soil-moisture sensing errors. This work provides a reproducible simulation framework for component-level (attributional) green value evaluation and for optimizing promotion strategies of smart agriculture platforms.
\end{abstract}


\section{Introduction}

The policy agenda of agricultural green transition (fertilizer and pesticide reduction, non-point source pollution control, dual-carbon goals) together with the rapid deployment of digital agriculture technologies makes ``how much green benefit can a smart agriculture platform deliver'' a question of common interest to academia and industry \cite{moa2024,ccg2025}. Existing studies mostly evaluate single technologies (soil testing \cite{zhang2018}, integrated water-fertilizer management \cite{moa2013}, paddy water management \cite{IPCC2019,lindau1991}) or the platform as a whole \cite{saltelli2008,lam2014}; studies that decouple AI components (intelligent diagnosis, question-answering push, remote-sensing identification) from perception-and-execution components (IoT sensing, smart irrigation) and attribute benefits to each are still rare. This gap leaves platform resource allocation without evidence: should we improve algorithm accuracy first, or farmer adoption first? Should we buy high-accuracy sensors, or invest in the AI decision module?

Using three representative tropical agriculture scenarios in Hainan as background, this paper builds on a previous whole-platform simulation framework, makes the components explicit, and designs two controlled Monte Carlo experiments to answer these questions. Experiment 1 focuses on the chemical input (pesticide, fertilizer) chain; Experiment 2 focuses on the irrigation water and paddy methane chain. Both experiments share fixed random seeds and the parameter source-grading system (A = official/methodology, B = literature, C = expert intervals), and use the previous whole-platform results as calibration anchors to keep the research chain consistent.

\section{Materials and Methods}

\subsection{Overall Design}

The platform's green value is decomposed into two orthogonal chains:

\begin{align}
\text{Chain I (green behavior):}\quad & \text{reduction}= \underbrace{\varphi_{avoid}}_{\text{avoidable}}\times \underbrace{\eta}_{\text{effectiveness}}\times \underbrace{\gamma}_{\text{coverage}}\times \underbrace{\alpha}_{\text{adoption}}\\
\text{Chain II (sense-execute):}\quad & \text{saving}= \underbrace{C}_{\text{tech ceiling}}\times \alpha \times \underbrace{q}_{\text{scheduling quality}}\times (1-\kappa\,\sigma_{IoT})
\end{align}

Experiment 1 compares $\gamma=\gamma_{exp}$ (extension mode) with $\gamma=\gamma_{AI}$ (AI mode), with $\eta_{AI}=\eta(acc)$ depending on diagnostic accuracy; Experiment 2 compares an engineering stage and a decision stage (P1: IoT engineering + human dashboard reading; P2: +AI optimal scheduling). Scenario scopes (Hainan mango orchards, winter vegetables, rice/nanfan; area weights 40\%:30\%:30\%) and carbon baselines (cradle-to-farm-gate) follow the previous study; parameters are listed in Tables~\ref{tab:p1} and~\ref{tab:p2}. All uncertainties use triangular distributions $\mathrm{Tri}(lo, mode, hi)$; each Monte Carlo experiment iterates 10,000 times with seeds 20260907 (Exp.~1) and 20260908 (Exp.~2), fully reproducible. Parameter grading follows the A/B/C system: A = official documents or international methodologies \cite{moa2013,IPCC2019}; B = peer-reviewed literature; C = expert-judgment intervals, to be updated by Bayesian methods after pilot measurements.

\begin{table}[htbp]
\centering
\caption{Experiment 1 parameters (triangular lo/mode/hi)}\label{tab:p1}
\small
\begin{threeparttable}
\setlength{\tabcolsep}{3.5pt}
\begin{tabular}{llclll}
\toprule
Parameter & Meaning & lo & mode & hi & Grade\\
\midrule
$\varphi_{avoid}$ & avoidable blind sprays share & 0.15 & 0.30 & 0.50 & C\\
$\eta_{exp}$ & extension prescription effectiveness & 0.55 & 0.70 & 0.85 & C\\
$\gamma_{exp}$ & extension coverage & 0.30 & 0.40 & 0.55 & C\\
$\gamma_{AI}$ & AI coverage & 0.85 & 0.92 & 0.97 & B/C\\
$acc$ & AI diagnostic accuracy & 0.78 & 0.88 & 0.95 & B\\
$\alpha$ & AI suggestion adoption rate & 0.35 & 0.60 & 0.85 & C\\
$\varphi_{soil}$ & soil-test reduction & 0.05 & 0.15 & 0.30 & B\\
$\varphi_{org}$ & organic substitution & 0.10 & 0.30 & 0.55 & B\\
$\gamma_{f,exp}$ & extension prescription-to-door rate & 0.15 & 0.25 & 0.40 & C\\
$\gamma_{o,exp}$ & extension organic-to-door rate & 0.20 & 0.35 & 0.55 & C\\
\bottomrule
\end{tabular}
\begin{tablenotes}\scriptsize
\item[Note] AI prescription effectiveness follows $\eta_{AI}=0.70+0.25\,(acc-0.70)$; fertilizer reduction is the prescription term $0.7\varphi_{soil}+0.3\varphi_{org}$ discounted by the to-door rates.
\end{tablenotes}
\end{threeparttable}
\end{table}

\begin{table}[htbp]
\centering
\caption{Experiment 2 parameters (lo/mode/hi, three stages)}\label{tab:p2}
\small
\begin{threeparttable}
\setlength{\tabcolsep}{3pt}
\begin{tabular}{llclll}
\toprule
Parameter & Meaning & lo & mode & hi & Grade\\
\midrule
$C_{m}/C_{v}/C_{r}$ & tech ceiling mango/veg/rice & 0.27/0.22/0.13 & 0.40/0.35/0.23 & 0.55/0.50/0.35 & A/B\\
$q$ (P0/P1/P2) & scheduling quality & 0.30/0.50/0.80 & 0.42/0.62/0.88 & 0.55/0.75/0.95 & C\\
$\alpha$ & technology adoption rate & 0.30 & 0.55 & 0.80 & C\\
$\sigma_{IoT}$ & soil-moisture observation error & 0.02 & 0.04 & 0.07 & B\\
$\kappa$ (P1/P2) & error sensitivity & 1.0 & --- & 0.4 & C\\
$\psi_{CH4}$ & AWD potential & 0.20 & 0.35 & 0.55 & B\\
$e_{CH4}$ (P1/P2) & CH$_4$ execution & 0.40/0.75 & 0.55/0.85 & 0.70/0.93 & C\\
\bottomrule
\end{tabular}
\begin{tablenotes}\scriptsize
\item[Note] P2 parameters are calibrated against the previous whole-platform model: median aggregate saving 16.0\% (vs.\ 16.5\%), CH$_4$ reduction 30.5\% (vs.\ 30.0\%), deviation $<$4\%.
\end{tablenotes}
\end{threeparttable}
\end{table}

\section{Results and Analysis}

\subsection{Experiment 1: Marginal Contribution of AI to Pesticide--Fertilizer Reduction}

\begin{figure}[htbp]
\centering
\includegraphics[width=0.95\textwidth]{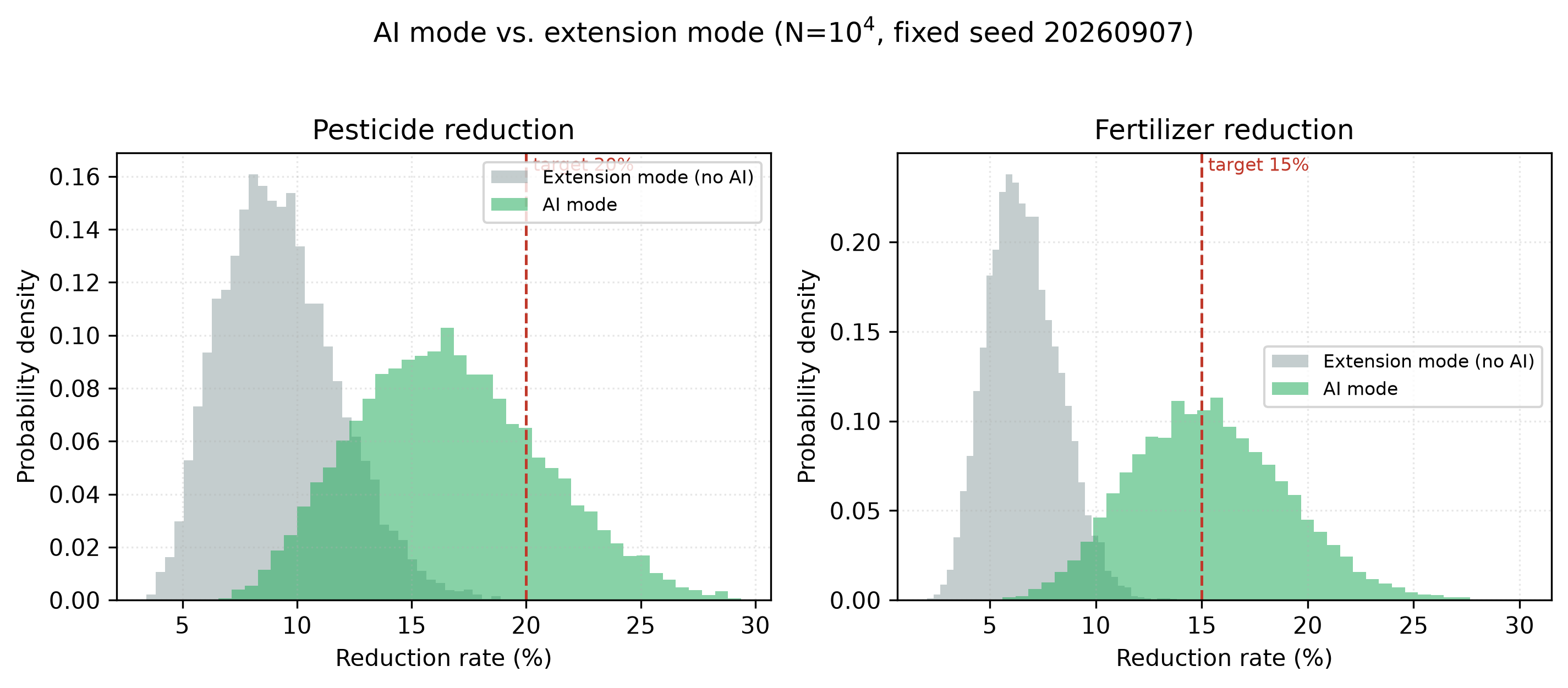}
\caption{Distribution comparison of extension vs.\ AI modes (left: pesticide; right: fertilizer)}
\label{fig:e1dist}
\end{figure}

With 10,000 iterations (Table~\ref{tab:r1}): the median pesticide reduction in the extension mode is only 9.0\% and fertilizer reduction 6.5\%; under the AI mode these rise to 16.4\% and 15.2\%, with AI median increments of 7.1 and 8.5 percentage points, respectively (Figure~\ref{fig:e1dist}). Attainment probabilities: in the extension mode, both ``pesticide $\geq$20\%'' and ``fertilizer $\geq$15\%'' are essentially zero; under AI they rise to 20.7\% and 52.0\%.

\begin{table}[htbp]
\centering
\caption{Experiment 1 Monte Carlo results (N=10$^{4}$, seed 20260907)}\label{tab:r1}
\small
\begin{threeparttable}
\begin{tabular}{lcccc}
\toprule
Variable & Mean & Median & P5 & P95\\
\midrule
Pesticide reduction (extension) & 0.093 & 0.090 & 0.055 & 0.138\\
Pesticide reduction (AI) & 0.166 & 0.164 & 0.104 & 0.237\\
Fertilizer reduction (extension) & 0.066 & 0.065 & 0.040 & 0.095\\
Fertilizer reduction (AI) & 0.154 & 0.152 & 0.097 & 0.216\\
AI pesticide increment & 0.074 & 0.071 & 0.041 & 0.115\\
AI fertilizer increment & 0.088 & 0.085 & 0.051 & 0.133\\
\bottomrule
\end{tabular}
\end{threeparttable}
\end{table}

\begin{figure}[htbp]
\centering
\includegraphics[width=0.72\textwidth]{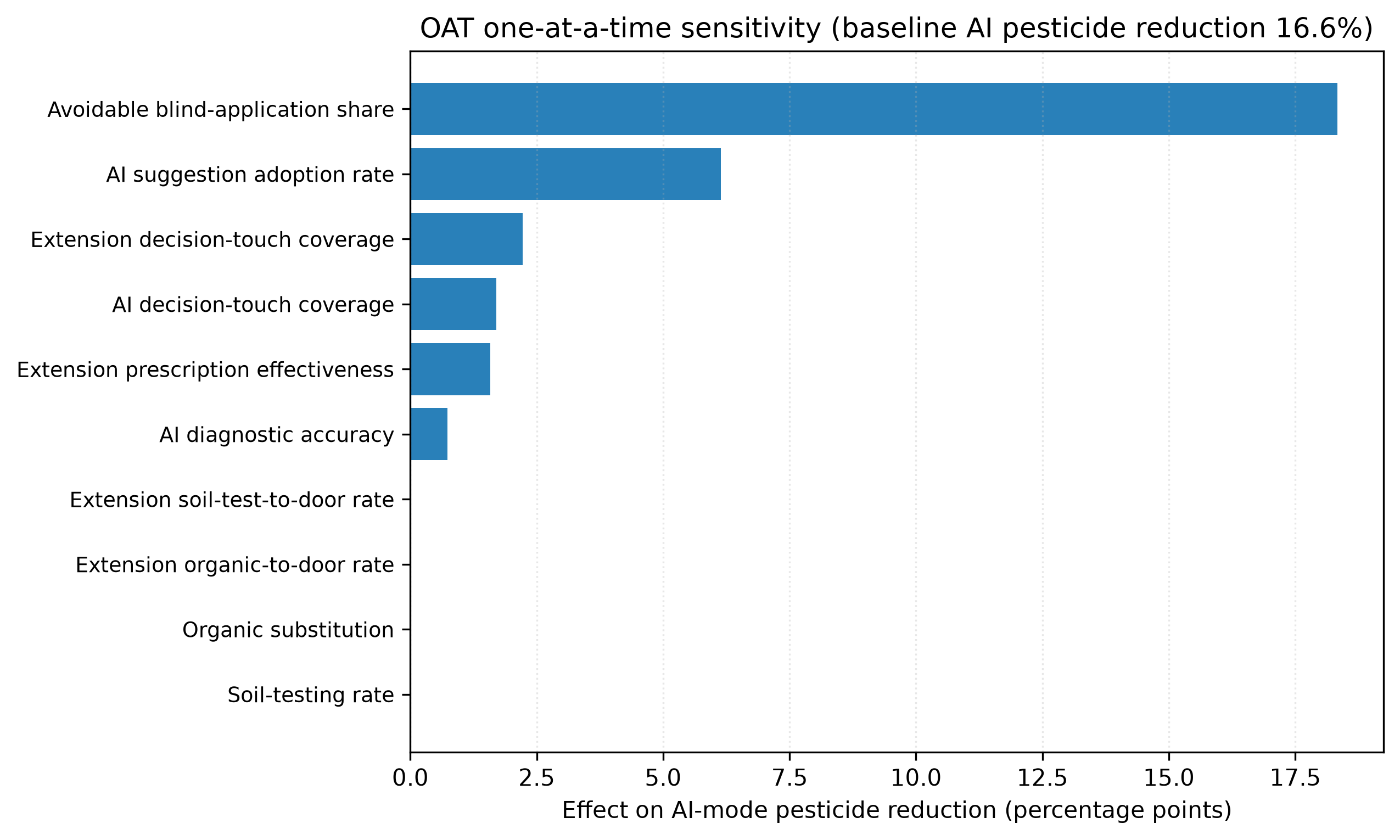}
\caption{Experiment 1 OAT sensitivity: effects on AI-mode pesticide reduction}
\label{fig:e1oat}
\end{figure}

The scenario matrix (Table~\ref{tab:matrix}) shows that raising adoption from 0.35 to 0.85 increases pesticide reduction by 5.5--6.4 pp, while raising accuracy from 0.70 to 0.95 adds only 0.5--1.5 pp; ``pesticide $\geq$20\%'' requires both accuracy and adoption at high levels (P = 49\% at 0.95$\times$0.85). The OAT sensitivity analysis (Figure~\ref{fig:e1oat}) likewise shows that after $\varphi_{avoid}$ (scenario endowment, uncontrollable), $\alpha$ (adoption) is the first controllable lever, while $acc$ has the smallest marginal effect.

\begin{table}[htbp]
\centering
\caption{Experiment 1 scenario matrix: AI pesticide reduction mean / P(pesticide $\geq$20\%)}\label{tab:matrix}
\small
\begin{threeparttable}
\begin{tabular}{lccc}
\toprule
$acc$ $\backslash$ $\alpha$ & 0.35 & 0.60 & 0.85\\
\midrule
0.70 & 13.1\% / 1\% & 15.9\% / 15\% & 18.6\% / 37\%\\
0.85 & 13.5\% / 2\% & 16.5\% / 20\% & 19.4\% / 43\%\\
0.95 & 13.7\% / 3\% & 16.9\% / 23\% & \textbf{20.1\% / 49\%}\\
\bottomrule
\end{tabular}
\end{threeparttable}
\end{table}

\subsection{Experiment 2: IoT and AI Contributions to Water Saving and Paddy Methane}

\begin{figure}[htbp]
\centering
\includegraphics[width=0.95\textwidth]{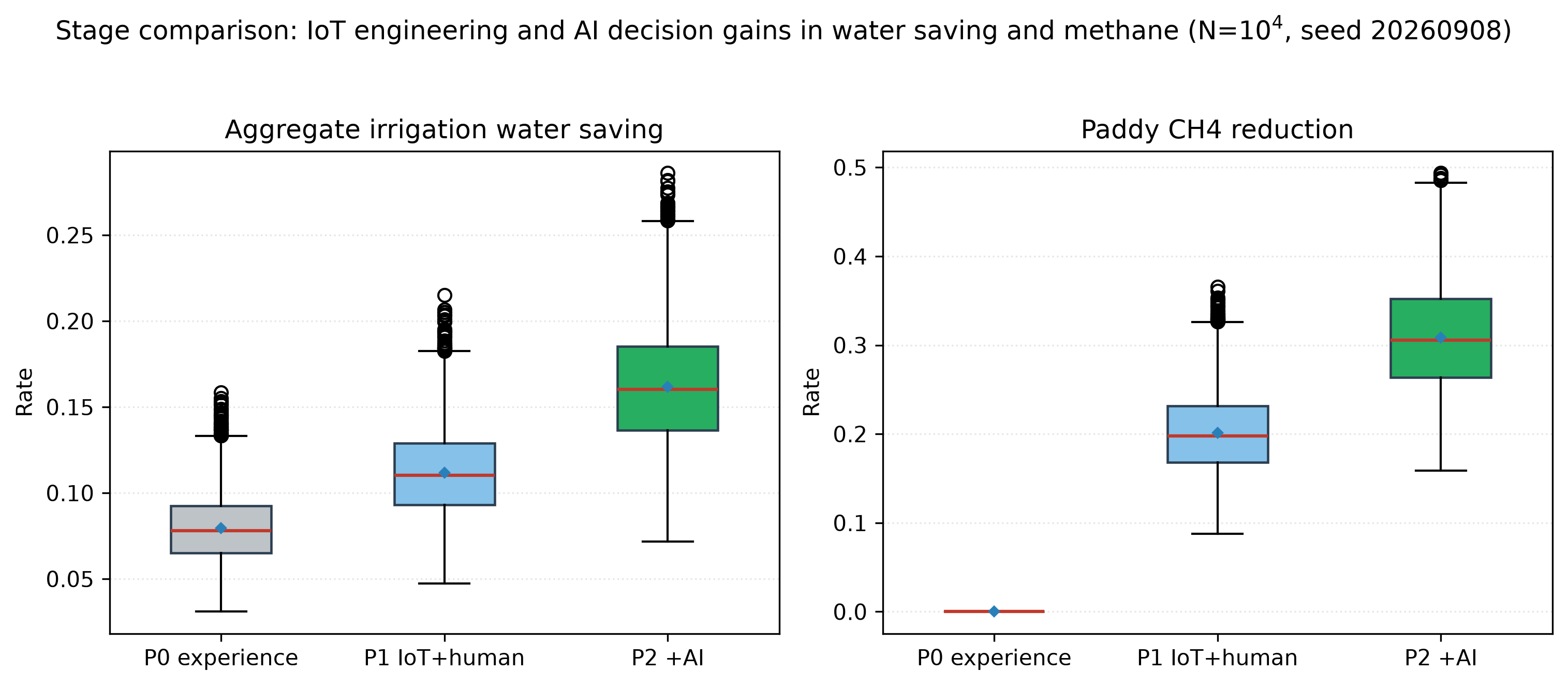}
\caption{Boxplots of three-stage aggregate water saving and paddy CH$_4$ reduction}
\label{fig:e2box}
\end{figure}

\begin{figure}[htbp]
\centering
\includegraphics[width=0.74\textwidth]{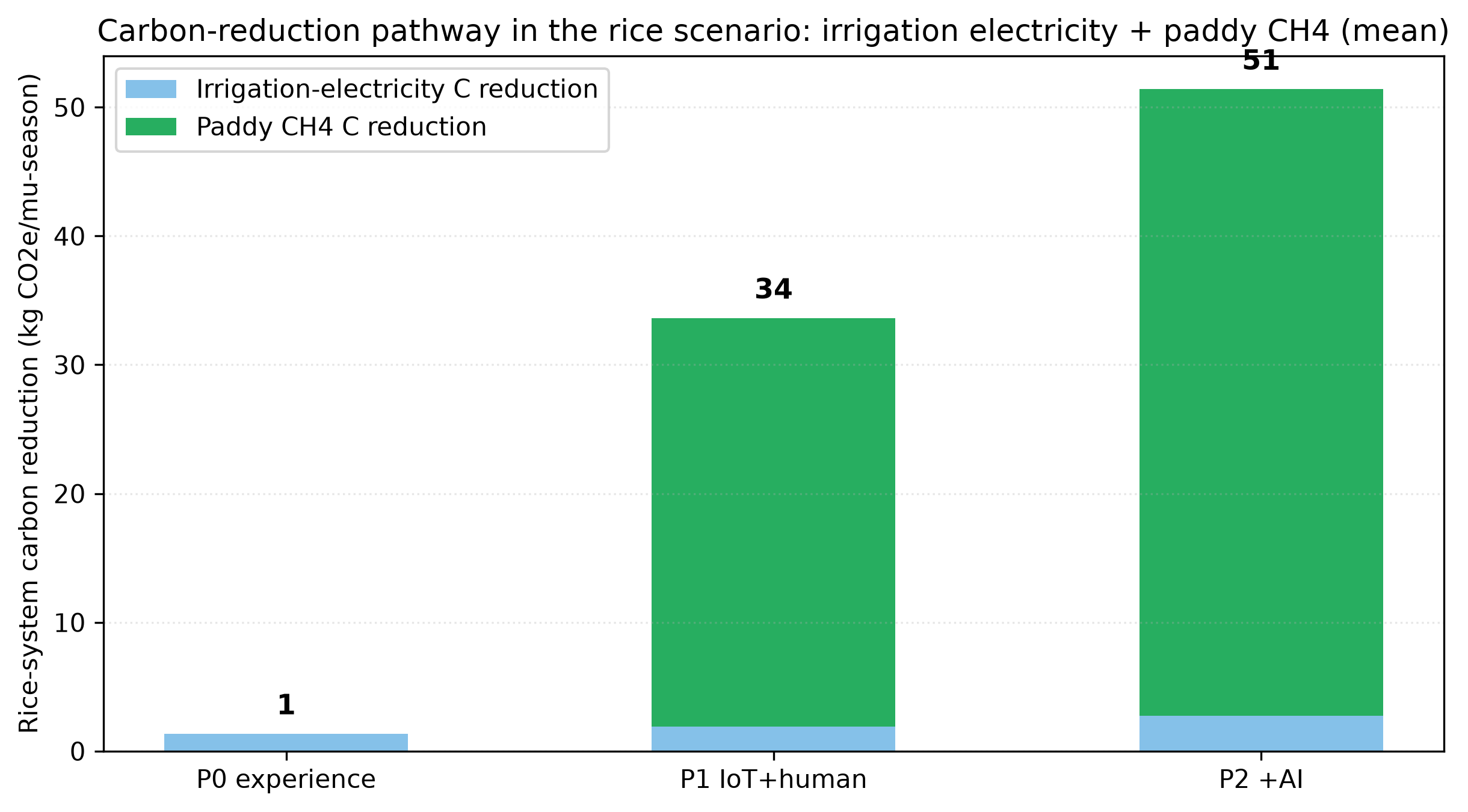}
\caption{Carbon-reduction pathway decomposition in the rice scenario: irrigation electricity and paddy CH$_4$ (means, kg CO$_2$e/mu-season)}
\label{fig:e2path}
\end{figure}

Three-stage comparison (Table~\ref{tab:r2}, Figure~\ref{fig:e2box}): median aggregate water saving rises from 7.8\% (P0) to 11.0\% (P1) and then 16.0\% (P2); the marginal contribution of AI decisions over engineering retrofit is 5.0 pp. The probability of ``aggregate saving $\geq$15\%'' rises from 0.1\% (P0) to 7.9\% (P1) and 60.9\% (P2), while ``$\geq$20\%'' remains only 14.4\% at baseline adoption. Paddy CH$_4$ reduction is only 19.8\% under P1 manual execution but reaches 30.5\% under P2 AI scheduling; the carbon intensity of the rice ``irrigation--methane'' subsystem declines 27.9\%, with the paddy CH$_4$ pathway contributing about 95\% (Figure~\ref{fig:e2path}).

\begin{table}[htbp]
\centering
\caption{Experiment 2 three-stage results (medians)}\label{tab:r2}
\small
\begin{threeparttable}
\begin{tabular}{lcccc}
\toprule
Indicator (median) & P0 current & P1 IoT engineering & P2 +AI decision\\
\midrule
Mango water saving & 0.092 & 0.131 & 0.190\\
Vegetable water saving & 0.081 & 0.114 & 0.166\\
Rice water saving & 0.054 & 0.076 & 0.110\\
Aggregate water saving & 0.078 & 0.110 & 0.160\\
Paddy CH$_4$ reduction & 0.000 & 0.198 & 0.305\\
Rice subsystem carbon reduction & 0.007 & 0.182 & 0.279\\
\bottomrule
\end{tabular}
\end{threeparttable}
\end{table}

\begin{figure}[htbp]
\centering
\includegraphics[width=0.68\textwidth]{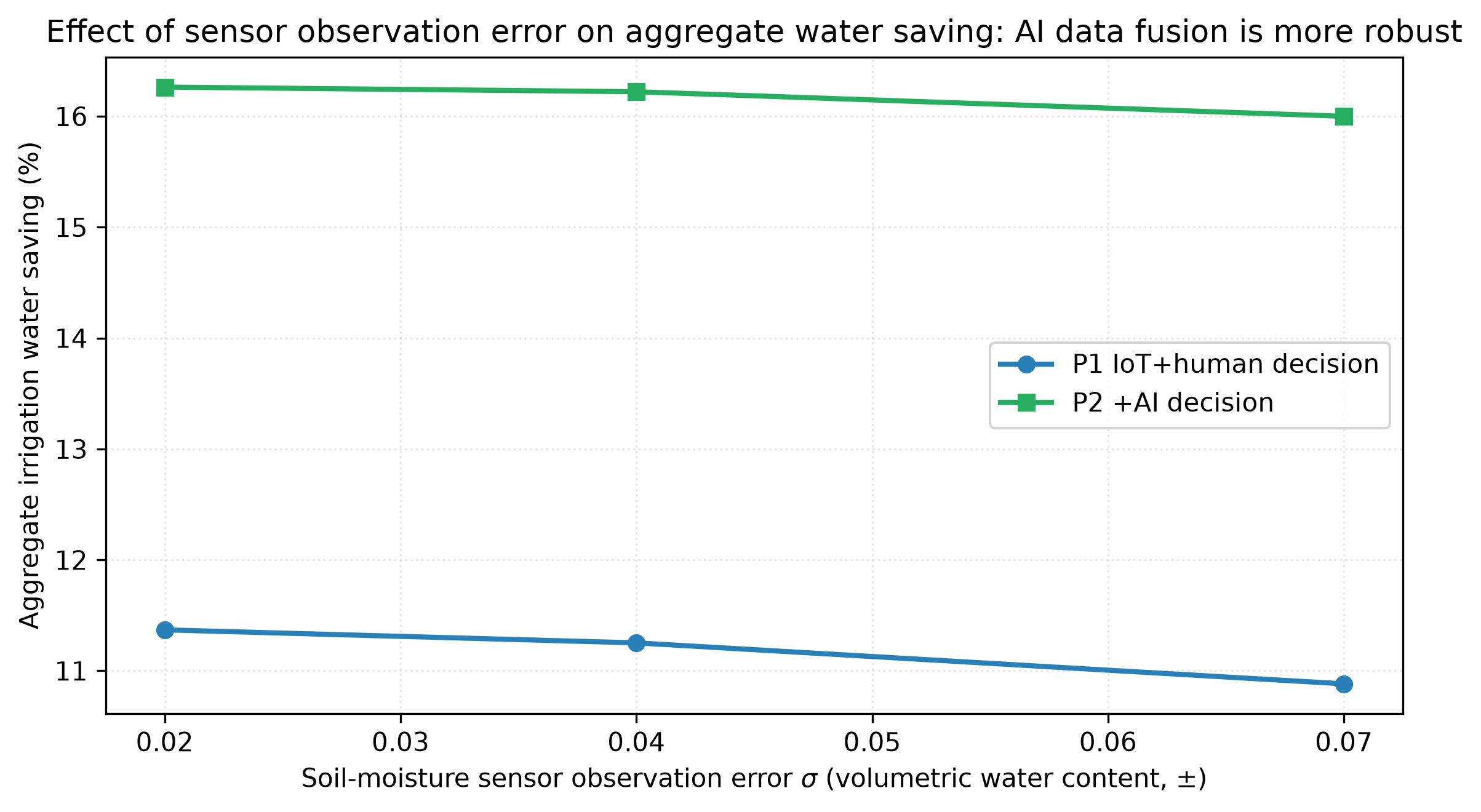}
\caption{Effect of soil-moisture observation error on aggregate water saving: AI decision is more robust}
\label{fig:e2sigma}
\end{figure}

Sensor-error scenarios (Figure~\ref{fig:e2sigma}): as $\sigma$ rises from 0.02 to 0.07, P1 loses 0.5 pp of aggregate saving and 3.4 pp of attainment probability, whereas P2 loses only 0.3 pp and 2.3 pp; AI data fusion is robust to sensing noise and relaxes hardware precision constraints. Adoption-threshold analysis shows that at P2 adoption $\geq$0.85 the median aggregate saving reaches 24.8\% with P($\geq$20\%) $\approx$ 98\%, i.e., the ``20\% water saving'' target is attainable under AI mode with high adoption.

\subsection{Reconciliation with the Previous Whole-Platform Study}

Deviations of P2 (full function) medians from the corresponding medians of the previous whole-platform Monte Carlo study \cite{saltelli2008} are: mango 3.6\%, vegetable 1.8\%, rice 2.0\%, aggregate 2.9\%, CH$_4$ reduction 1.8\%, all below 4\%, confirming that component-wise decomposition does not change the overall magnitude. Under full adoption ($\alpha=1$), the Experiment 1 pesticide and fertilizer reductions are also consistent with the whole-platform scope (about 22\% and 19.5--21\%). The two studies thus form a closed ``whole $\rightarrow$ components'' chain of evidence.

\section{Discussion}

\subsection{Mechanism Decomposition of AI's Green Value}

The results show that AI's green value = touch-amplifier ($\gamma_{exp}$ 0.40 $\rightarrow$ $\gamma_{AI}$ 0.92) $\times$ quality-amplifier ($\eta(acc)$), with the former dominant. This is consistent with the practical observation that digital-agriculture promotion fails at farmer reach rather than at the prescription itself \cite{ccg2025}, and quantitatively explains the diminishing marginal returns of pure model-accuracy improvements: in the 80--95\% range, reach and adoption are the binding constraints.

\subsection{Implications for Platform Operation and Green Investment}

(1) Resource priority should be: farmer adoption operations $>$ sensing and execution infrastructure $>$ algorithm accuracy optimization; the sensitivity rankings of the two experiments corroborate each other. (2) The ``water saving $\geq$20\%'' target is attainable under AI with high adoption (P2 $\times$ $\alpha$ 0.85: P $\approx$ 98\%), providing a feasible path for the earlier conservative conclusion, but must be qualified by an adoption-rate threshold statement. (3) AWD-based CH$_4$ abatement is the dominant carbon-reduction pathway in the rice scenario (about 95\% of the subsystem), and AI scheduling can raise execution from 0.55 (manual) to 0.85, a low-cost high-benefit green lever. (4) AI's robustness to sensor noise implies that medium/low-accuracy sensors combined with AI correction can lower the hardware investment threshold.

\subsection{Limitations and Outlook}

This study remains an ex-ante simulation: diagnostic accuracy, adoption and execution parameters come mainly from literature and expert ranges and await pilot calibration (first-wave observations suggested: expert-review accuracy of AI diagnoses, farmer adoption logs, gate execution logs, water meters and application ledgers). Learning curves, seasonal dynamics and the yield risk of AI misdiagnosis are not modeled; the sensor model excludes failure rates and transmission packet loss. Future work will update parameters by Bayesian methods with pilot data and extend the framework to nutrient management and post-harvest stages.

\section{Conclusions}

(1) The marginal contribution of AI components to pesticide--fertilizer reduction is substantial: the extension mode can hardly reach ``pesticide $\geq$20\% / fertilizer $\geq$15\%'' (P $\approx$ 0), while the AI mode reaches 20.7\% and 52.0\% (baseline), with a median fertilizer increment of 8.5 pp.

(2) In the water chain, IoT engineering (P1) contributes 3.2 pp of aggregate saving and AI decisions (P2) add 5.0 pp more; paddy CH$_4$ reduction reaches 30.5\% under AI scheduling and the rice irrigation--methane subsystem carbon intensity declines 27.9\%.

(3) The primary bottleneck for green-target attainment is farmer adoption rather than algorithm accuracy; AI data fusion is robust to sensing error. Platform operation should prioritize adoption and trust.

(4) Component-wise Monte Carlo experiments deviate from the whole-platform model by less than 4\%, forming a closed ``whole $\rightarrow$ components'' evidence chain and providing a methodological framework for attributable green-value evaluation of smart agriculture platforms.

\section*{Data and Code Availability}
All scripts and sample data of both experiments (fixed seeds 20260907/20260908, N=10$^{4}$, one-command reproducible) are in the delivery directories scripts/ and output/.

\section*{Acknowledgements}
The authors thank the Sanya University innovation and entrepreneurship incubation platform, Handan Weizhi Jingjie AI Basic Software Co., Ltd., and the agricultural expert advisory team for their support in parameter-range definition and knowledge-base construction. This work was supported by the National College Student Innovation and Entrepreneurship Training Program (China).

\renewcommand{\refname}{References}
\bibliography{refs2}

@article{moa2024,
  author  = {{农业农村部}},
  title   = {全国智慧农业行动计划（2024--2028年）},
  journal = {农业农村部公报},
  year    = {2024}
}

@article{ccg2025,
  author  = {{中共中央、国务院}},
  title   = {乡村全面振兴规划（2024--2027年）},
  journal = {国务院公报},
  year    = {2025}
}

@article{moa2013,
  author  = {{农业部办公厅}},
  title   = {水肥一体化技术指导意见},
  journal = {农业部办公厅文件（农办农〔2013〕3号）},
  year    = {2013}
}

@techreport{IPCC2019,
  author      = {{IPCC}},
  title       = {2019 Refinement to the 2006 IPCC Guidelines for National Greenhouse Gas Inventories},
  institution = {IPCC},
  address     = {Geneva},
  year        = {2019}
}

@article{zhang2018,
  author  = {张福锁 and 吴文良 and others},
  title   = {测土配方施肥与化肥减量增效：研究与实践进展},
  journal = {中国农业科学},
  year    = {2018},
  volume  = {51},
  number  = {14},
  pages   = {2651--2662}
}

@article{lindau1991,
  author  = {Lindau, C. W. and Bollich, P. K. and DeLaune, R. D. and others},
  title   = {Effect of urea fertilizer and environmental factors on CH$_4$ emissions from a Louisiana, USA rice field},
  journal = {Plant and Soil},
  year    = {1991},
  volume  = {136},
  pages   = {195--203}
}

@article{lam2014,
  author  = {Lam, S. K. and Chen, D. and Norton, R. and Armstrong, R. and Mosier, A. R.},
  title   = {Influence of climate change, soil carbon and fertilizer management on nitrogen cycling: a research synthesis},
  journal = {Global Change Biology},
  year    = {2014},
  volume  = {20},
  number  = {1},
  pages   = {253--266}
}

@book{saltelli2008,
  author    = {Saltelli, A. and Ratto, M. and Andres, T. and Campolongo, F. and Cariboni, J. and Gatelli, D. and Saisana, M. and Tarantola, S.},
  title     = {Global Sensitivity Analysis: The Primer},
  publisher = {John Wiley \& Sons},
  address   = {Chichester},
  year      = {2008}
}

\end{document}